\def\year{2020}\relax
\documentclass[letterpaper]{article} 
\usepackage{aaai20}  
\usepackage{times}  
\usepackage{helvet} 
\usepackage{courier}  
\usepackage[hyphens]{url}  
\usepackage{graphicx} 
\usepackage{graphicx}  
\usepackage{amsmath,amsfonts,bm}

\def\eqref#1{equation~\ref{#1}}

\def\1{\bm{1}}
\newcommand{\train}{\mathcal{D}}

\DeclareMathAlphabet{\mathsfit}{\encodingdefault}{\sfdefault}{m}{sl}
\SetMathAlphabet{\mathsfit}{bold}{\encodingdefault}{\sfdefault}{bx}{n}

\makeatletter
\newcommand{\removelatexerror}{\let\@latex@error\@gobble}
\makeatother

\usepackage{url}
\usepackage{amsmath,amsfonts,amssymb}
\usepackage[ruled,norelsize,linesnumbered]{algorithm2e}
\usepackage{algorithmic}
\usepackage{tabulary} 
\usepackage{rotating} 
\usepackage{multirow} 
\usepackage{booktabs}

\newcommand{\expnumber}[2]{{#1}\mathrm{e}{#2}}

\usepackage{color}
\usepackage{soul}

\title{Single-Agent Optimization Through Policy Iteration Using Monte-Carlo Tree Search}

\author{Arta Seify \textnormal{and} Michael Buro\\
University of Alberta\\
Edmonton, Canada\\
\{seify,mburo\}@ualberta.ca}

\begin{document}
	
	\maketitle
	
	\begin{abstract}
		The combination of Monte-Carlo Tree Search (MCTS) and deep reinforcement
learning is state-of-the-art in two-player perfect-information games. In this
paper, we describe a search algorithm that uses a variant of MCTS which we
enhanced by 1) a novel action value normalization mechanism for games with
potentially unbounded rewards (which is the case in many optimization
problems), 2) defining a virtual loss function that enables effective
search parallelization, and 3) a policy network, trained by generations of self-play, 
to guide the search.
We gauge the effectiveness of our method in ``SameGame''---a popular
single-player test domain. Our experimental results indicate that our method
outperforms baseline algorithms on several board sizes. Additionally, it is
competitive with state-of-the-art search algorithms on a public set of positions.

	\end{abstract}
	
	
	\section{Introduction}
Single-agent optimization problems have been an active field of research for
decades. Such problems include any domain with an agent whose goal is to
maximize an objective function(s), without interference from any other agents.
NP-hard problems such as the Travelling Salesman Problem (TSP) can be
framed as a single-agent optimization problem. Algorithms for solving TSP have
many practical uses, such as computer chip design and order-picking in
warehouses \cite{theys2010using}. Single-agent optimization problems can be
represented as (deterministic) single-player games. This is the term used
throughout this paper to better present our contribution in relation to
previous work.

Most state-of-the-art heuristic search algorithms for single-player games use
Monte-Carlo simulations \cite{schadd2012single}, \cite{cazenave2009nested}, \cite{rosin2011nested}. These methods estimate the
values of states using random simulations. The generality of these methods makes them
applicable to a wider variety of domains. Examples include two-player board games
such as Go
\cite{gelly2012grand} and Hex \cite{arneson2010monte}, \cite{huang2013mohex},
real-time domains such as Ms. Pac-Man \cite{pepels2014real}, general video
game playing \cite{perez2019general}.

A recently proposed enhancement is to combine MCTS with deep reinforcement learning \cite{silver2018general},
\cite{anthony2017thinking}. These algorithms have achieved state-of-the-art
performance in deterministic two-player perfect-information zero-sum
games. MCTS provides the agent with the ability to look ahead, while policy
and value networks are used to decrease the width and depth of the search
tree, respectively. Additionally, the trained policy
networks can be surprisingly strong by themselves. For instance, AlphaZero's
Go policy plays at human expert level without the need for forward search.

In this paper, our goal is to bring the ideas of these algorithms to
single-player optimization problems. There are multiple differences between
the two settings that make this task non-trivial. In zero-sum two-player
games, the reward seen by the agent is often one of $\{-1, 0, 1\}$, for loss,
tie, or win, whereas in single-player games, the reward are unknown. 
Furthermore, values found during search are a lower-bound on
the optimal value. Therefore, the action selection strategy, as well as the
policy target, have to be adjusted.

To address these problems, after a discussion of related work, we first
introduce our policy-guided MCTS algorithm for single-agent optimization
problems, then describe how we train our policy networks, and finally measure
the performance of our methods in the NP-hard SameGame---a popular single-agent
optimization domain---before closing with concluding remarks and suggestions
for future work.


	\section{Related Work} \label{sec:relatedwork}

We begin by introducing several state-of-the-art search algorithms for
single-player games. Then, pioneer works on combining MCTS and deep
reinforcement learning for two-player games are presented. Lastly, we discuss
similar work on single-player games.

\subsection{MCTS in Single-Player Games}

Schadd \textit{et al.} introduced Single-Player MCTS
(SP-MCTS) \cite{schadd2012single}, which was the first successful application of MCTS to a
single-player game with a large state and action space. Two contributions which we also use in our work are 
creating a tree-per-move rather than tree-per-game by playing the highest
valued action, and a few value normalization methods.

Cazenave developed a recursive variant of MCTS called Nested Monte-Carlo
Search (NMCS) \cite{cazenave2009nested}. NMCS estimates the values of states
at level $k$ using a recursive call of level $k-1$, where a level 1 recursion
is a Monte-Carlo rollout.

Rosin's Nested Rollout Policy Adaptation (NRPA) \cite{rosin2011nested}
combines NMCS with online policy learning. NRPA's rollouts are guided by a
policy, which is slowly adapted towards the moves with the highest return. An
efficient encoding of states, which transforms similar states to the same
value, is required for the algorithm to perform well.


\subsection{MCTS and Deep Reinforcement Learning}

The combination of MCTS and deep reinforcement learning has been successfully
applied to two-player games. One of the major contributions is the work on
AlphaZero \cite{silver2018general}, which reached superhuman performance in
the games of Go, Chess, and Shogi. The algorithm uses a single two-headed
network, which outputs both a policy and value of a state. The policy is used
during the selection step of MCTS, and the simulation phase is replaced by the
value prediction. The network is trained from scratch using self-play, without
the use of any human generated data.

A similar work to AlphaZero is Expert Iteration \cite{anthony2017thinking}.
Expert Iteration trains a new neural network at each generation, starting with
only a policy network, which is used in a similar way as
AlphaZero. Once the generated data is of sufficient
quality, the policy network is replaced by a two-headed network. The value prediction of the network is combined
with the rollout result using a mixing parameter, which further improves the agent.

We are not the first to combine MCTS with deep reinforcement learning for
single-player games. The work of \cite{mcaleer2018solving} trains a two-headed
network to solve the Rubik's cube, using a training procedure called
Autodidactic Iteration (ADI). ADI is only suitable for problems in which there is a
single known goal state. This is in contrast to
our algorithm, which---as we will see---does not have this requirement.

Laterre \textit{et al.} introduce Ranked-Rewards
(R2 \cite{laterre2018ranked}), a general algorithm that enables self-play for single-player games. This
is accomplished by setting the rewards seen by the agent to a value of either
$1$ or $-1$, depending on whether the actual value of the state is higher or
lower than a given percentile of previously seen rewards. This pits the agent
against itself, forcing it to continuously outperform previous
generations. Our algorithm does not make use of this self-play scheme. 
	\section{Policy-Guided MCTS for Single-Agent Optimization}

Our work is based on the belief that combining MCTS and deep reinforcement
learning can also lead to state-of-the-art performance in single-player games.
Therefore, our objectives are threefold: 1) develop an effective variant of
MCTS for single-agent domains, 2) establish a suitable learning target for the
policy network, and 3) create a training procedure similar to the self-play
regime for two-player games.

\subsection{Integrating Policies into MCTS} \label{sec:MCTS}

Our initial task is to develop a policy-guided MCTS algorithm for
single-player domains. To this end, we first introduce a normalization
strategy that is applied during the selection stage of MCTS. Having normalized
values allows us to use the standard Predictor + Upper Confidence Bound For
Trees (PUCT) \cite{silver2018general} selection strategy, which we introduce
in Eq.~\ref{eq:puct}. We additionally use the policy for the initial action
selection in expanded nodes, as well as to guide the rollouts. Using a neural
network during search is expensive however, so we define a virtual loss
function that enables tree parallelization
\cite{chaslot2008parallel}. Parallelization allows us to do batch prediction,
which increases GPU efficiency, thereby increasing the speed of the search.

\subsubsection{Value Normalization}

To use the PUCT selection strategy, the rewards need to be in the range of $[-1,
1]$. In single-player games however, the range of rewards is often different,
and unknown. Therefore, we require a robust method for normalizing the values.

Several normalization techniques have been introduced in previous work
\ifdefined\aamas \cite{schadd2012single, klein2015attacking} \else
\cite{schadd2012single}, \cite{klein2015attacking} \fi. In SP-MCTS, the values
are not normalized, instead larger parameters are used in the selection
strategy, which increases the upper confidence bound (UCB) term. The work of
\cite{klein2015attacking} uses the highest score achievable in any SameGame
board to normalize the values. The main problem with both strategies is that
they are domain specific, and require either extensive experiments or domain
knowledge to set properly.

We propose a more general normalization method---max-min scaling---that is
applied locally at the node level as follows:
\begin{equation}
\bar{Q}_{norm}(s, a) ~=~ \frac{2(Q(s, a) - min_{a'}Q(s,
	a'))}{max_{a'}Q(s, a') - min_{a'}Q(s, a')} - 1.
\end{equation}

\noindent
When no action has been taken in state $s$ (i.e., max and min values are not
yet defined), or they are equal, we set $Q_{norm}(s, a)$ to 1---being
optimistic. The highest and lowest values can be stored in a node directly, or
calculated by looping over all edges. This approach does not require any
assumptions about the lowest or highest achievable values in the
domain. Furthermore, it does not require the tracking of standard deviations
to relate average values and UCB exploration terms
\cite{schadd2012single}. Lastly, since $\bar{Q}$ is defined locally, action
values of one node do not impact other nodes, which is a desirable property.
  
Note that max-min normalization maximizes the spread of mapped action values:
we can expect at least one action with a value of 1, and after several
simulations, at least one action with a value of -1. Assuming a scenario where
the maximum and minimum values are very similar, and the reward bounds are
known, max-min normalization can lead to a higher level of exploitation than
standard UCT \cite{kocsis2006bandit}. One method for reducing the spread of
mapped action values is to use max-min normalization with the actual rewards,
rather than average state-action values. Since both methods normalize values
to range $[-1, 1]$, we can expect UCB-based selection strategies to converge
to the globally optimal solution.
  


\subsubsection{MCTS Stages}
In what follows, we describe the four stages of our MCTS algorithm that uses
max-min scaling, PUCT, and virtual loss enabling effective parallelization:

\textbf{1.~Selection.} The selection strategy is applied recursively until an
edge of a leaf node or an edge leading to a terminal node is selected. At each node,
the, the
action with the highest sum of value and upper confidence bound, corrected for
virtual loss, is selected:
\begin{equation}
a' = argmax_a\left(\left(Q(s, a) - L(s, a)\right)_{norm} + U(s, a)\right),
\end{equation}

\noindent
where the upper bound $U(s, a)$ is given by PUCT, which is calculated as
follows:
\begin{equation}
U(s, a) = c_{puct} \pi_\theta(s, a) \frac{\sqrt{N(s)}}{1 + N(s, a)}. \label{eq:puct}
\end{equation}

This selection strategy is initially focused on actions with high prior
probability and low visit count, but asymptotically prefers actions with high
values.
When an edge is selected, the virtual loss count (which is described below) and
visit count are increased: $ W(s, a) \leftarrow W(s, a) + 1,~~N(s, a) \leftarrow N(s, a) + 1.$

\textbf{2.~Expansion.} A leaf node $n_L$ is expanded on the first visit and
added to the tree. All child edges, one per legal action $a$, are created and
initialized to $\{N(s_L, a) =$ $W(s_L, a) =$ $\bar{Q}(s_L, a) =$
$Q_{total}(s_L, a) =$
$0\}$, where $s_L$ is
the state of node $n_L$. The prior probability is also stored in the edge,
which is the re-normalized output of the network after filtering out illegal actions. 
We lock the node
and put the thread to sleep until the network is finished evaluating the
state.

\textbf{3.~Simulation.} The edge with the highest prior probability is selected from the newly expanded node $n_L$ as the first
action in the simulation. The rest of the simulation is either uniformly
random among all valid actions, or is guided by the current policy. The
policy-based rollout selects an action in state $s$ by sampling from the
policy. The simulation is finished once a terminal state is reached.

Right after the first rollout is finished, the $\bar{Q}$ value of all the
edges is set to the value of the simulation. This gives the other edges the
opportunity to be selected independently of the success of the first
simulation. This initialization strategy is ``optimistic'', as the 
greedy action is selected in a newly 
expanded node . Another option for optimistic initialization is to give edges that
have not yet been tried a normalized value of 1.

The policy-based rollout is far more informed, but is much slower than the
random counterpart, since a network prediction is required at each step of
the simulation. We use policy guided simulations for training, and compare the two simulation
strategies when testing the strength of the final policy.

\textbf{4.~Backpropagation.} The result $R$ of the rollout is propagated to
the root, updating edge statistics along the way as follows: $Q_{total}(s,a) \leftarrow Q_{total}(s,a) + R, \bar{Q}(s, a) \leftarrow Q_{total}(s,a)/{N(s,a)}, W(s,a) \leftarrow W(s,a) - 1.$

\subsubsection{Tree Parallelization and Virtual Loss}

The speed of Policy-MCTS is considerably lower than plain MCTS, since it
requires a prediction from the policy network at each expanded node, and
potentially at every step of the rollout. The slowdown caused by the network
can be reduced by using batch predictions, which requires a parallel version
of the algorithm. Common parallelization strategies for MCTS are root, leaf,
and tree parallelization \cite{chaslot2008parallel}. In our implementation, we
use tree parallelization with node and edge mutexes, and virtual loss.

In our tree parallelization, all search threads work on the same tree, with
mutexes used to avoid data corruption. However, given that both UCT and PUCT
are deterministic, we can expect the majority of threads to take similar paths
down the search tree. To discourage this behaviour, we can add a temporary
\textit{virtual loss} to actions as they are selected
\cite{chaslot2008parallel}. In two-player games, virtual loss corresponds to
virtual rollouts that resulted in a loss. However, in our setting, we have a 
score with unknown bounds, so we have to define what a ``loss'' means. We define
virtual loss as $L(s, a) = wW(s, a)|\bar{Q}(s, a)|$,
where $W(s, a)$ is the virtual loss count stored in the edges of the tree, and $w$ is the global virtual loss weight, which is subject to optimization. 
Therefore, the loss is relative to the current state-action value. A benefit
of this approach is that it requires no knowledge about the bound of rewards. 

We optimized $w$ with respect to the final average score obtained. The
experiments used plain MCTS on 100 randomly generated 15 $\times$ 15 boards,
with 5 runs per board. We discovered that a value of 0.01 provided the highest
search speed, and was within a few points of the best average (not
statistically significant); we fixed $w=0.01$ in all of our experiments.

\subsection{Policy Training Target} \label{sec:policytarget}

In two-player adversarial games, assuming a limited search time budget for
MCTS, the edge with the highest visit count---as opposed to the highest valued
edge---is often selected as the action to play (e.g., \cite{silver2018general}
and \cite{anthony2017thinking}). Actions with higher visit counts are
considered more robust, as they guard against the case in which a newly
analyzed move with higher value, but much fewer simulations, is
overconfidently chosen.


By contrast, in single-player games, which are non-adversarial, the values of
simulations starting in a state are a lower bound on the maximum achievable
value. This means that the action with the highest simulation value is
currently the best action, regardless of how often other actions were
attempted. In the limit, the action with the highest visit count will also have
the highest value \cite{kocsis2006bandit}. However, this might not be the case
when given a limited search budget.

Given the above observation, we set the target for the policy as follows, with
ties between equal valued actions being broken randomly: $\pi(s_t, a) = 1$ if
$a = a_t$ and $0$, otherwise.


Using this policy target, training reduces to a supervised learning task that
seeks to minimize the average cross-entropy loss between the target policy
$\pi$ and its approximation $\pi_\theta$, for all training samples. I.e., for
a mini-batch of size $B$, its loss $L$ is given by: $L =
-\frac{1}{B}\sum_{i=1}^{B}\pi(s_i)^\mathsf{T} \log \pi_\theta(s_i)$.
  
\subsection{Data Generation}


In many single-player games, the initial actions have a large impact on the
final score the agent can obtain. An intuitive choice, then, is to let the
agent spend the majority, if not the entire MCTS budget, on determining the
best initial move. This would allow the agent to come up with the best first
move it possibly can. However, once the search budget is spent, the entire
sequence, and not just the first move, is reported. Since the entire search
budget has been spent, there is no opportunity to optimize the rest of these
moves further. It has been shown in \cite{schadd2012single}, as well as
confirmed by our own experiments, that committing to an action after a
fraction of the search budget is spent, thereby allowing the search to
optimize the remaining move sequence, can produce better results.

Committing to actions corresponds to playing a game; once a decision has been
made, it cannot be reversed. This allows MCTS to spend
more time in deeper
sections of the search tree, meaning actions near the middle and end of the
game receive more of the search budget than they would otherwise. Therefore,
we can expect the resulting action sequences to be better.

Inspired by this observation, our agent interleaves planning and playing. In
each state, the agent receives a constant planning budget of $k$ MCTS
simulations. Once the planning budget is spent, the best action, which is the one with the highest
value, is taken.
Although this forces the agent to commit to earlier actions, it opens up the
opportunity to better optimize subsequent moves. This process is repeated
until a terminal state is reached. Then, all state-action pairs taken to reach
the final state are returned, to be used as training data.


While training, we add Dirichlet noise to the prior probability of all actions
$a$ at the root of the tree --- using $(1-\epsilon)\pi_\theta(s_{root}) +
\epsilon \text{Dir}(\alpha)$ instead of $\pi_\theta(s_{root})$. $\text{Dir}(\alpha)$ is a random vector with
L$_1$-norm of 1. Using a low $\alpha$ value will add a high value of noise to
a few moves, whereas a higher value will add a more uniform amount of noise to
a larger number of moves. The added noise has the potential to increase the
UCB value of actions the current policy believes to be bad, which encourages
exploration.

The value of $\alpha$ was chosen experimentally. We used $\epsilon = 0.25$, as
proposed in \cite{silver2018general}, and tried a variety of $\alpha$ values
during training. These experiments were run for a few generations, 
and the $\alpha$ that produced the highest average value during
training was selected for the experiments presented in
the experiments section.
The value of $\alpha$ is dependent on the average number of
legal moves per game: for the 7 $\times$ 7 board, we tried the following
values \{0.5, 0.75, 1.0, 1.25\}, whereas for the 15 $\times$ 15 board we tried
\{0.15, 0.25, 0.4\}. From these, we extrapolated a value for the 10 $\times$ 10 board. 
These parameters are presented in Table~\ref{tab:trainparams}.

\subsection{Training Procedure}

\begin{figure}[b]
  \vspace{-0.4cm}
	\removelatexerror
        \small
	\begin{algorithm}[H]
		\LinesNumbered
		\DontPrintSemicolon
		\SetKwFunction{PMCTS}{Policy-MCTS}{}
		\SetKwFunction{generate}{Generate-Random-State}{}
		\SetKwFunction{newnetwork}{Initialize-New-Network}{}
		\SetKwFunction{split}{Split}{}
		\SetKwFunction{shuffle}{Shuffle}{}
		\SetKwFunction{append}{Append}{}
		\SetKwFunction{train}{Train}{}
		\SetKwFunction{earlystopping}{EarlyStopping}
		\SetKwFunction{push}{Push}{}
		\SetKwFunction{queue}{queue}{}
		\SetKwFunction{uniform}{Uniform-Over-Valid-Actions}
		\SetKwFunction{}{}
		\SetAlgoLined
		\textbf{Input}: \#generations $G$, \#runs
		$N$, training/validation buffer lengths $l_t, l_v$,
		training-validation split percentage $\lambda$ \;
		\textbf{Output}: Trained policy network \;
		
		$\pi_0 \sim \uniform{}$ \;
		$B_{training} = \queue{$\text{len}$=$l_t$}$\;
                $B_{validation} = \queue{$\text{len}$=$l_v$}$\;
		
		\For{$g = 1, ..., G$}
		{
			$B = [ ]$ \;
			\For{$r = 1, ..., N$}
			{
				$s \leftarrow \generate{}$ \;
				$[(s_1, a_1), ..., (s_{\scriptscriptstyle\boldsymbol{\text{T}}},
				a_{\scriptscriptstyle\boldsymbol{\text{T}}})]
				\leftarrow \PMCTS{$s, \pi_{g-1}$}$ \;
				$B \leftarrow B + [(s_1, a_1), ...,
				(s_{\scriptscriptstyle\boldsymbol{\text{T}}},
				a_{\scriptscriptstyle\boldsymbol{\text{T}}})]$
			}
			$\shuffle{$B$}$ \;
			$T, V \leftarrow \split{$B$, $ \lambda$}$ \;
			$B_{training} \leftarrow B_{training} + T$ \;
			$B_{validation} \leftarrow B_{validation} + V$ \;
			$\pi_{g} \leftarrow \pi_{\theta=\text{random}}$ \;
			$\train{$ \pi_{g}, B_{training}, B_{validation}$}$ \;
		}
		\textbf{Return}: $\pi_{G}$
		
		\caption{Policy training procedure}
		\label{alg:training}
	\end{algorithm}
\end{figure}

Our training procedure alternates between \textit{policy evaluation} and
\textit{policy improvement}, in a process known as \textit{policy iteration}.
Policy iteration is guaranteed to converge to an optimal policy in the tabular
case. This guarantee no longer holds in the function approximation case. In
spite of this, deep neural networks trained by policy iteration have surpassed
human players in multiple games \cite{silver2018general}, \cite{anthony2017thinking}.

Using MCTS in conjunction with the current policy to play complete games
constitutes the \textit{policy evaluation} step. Because MCTS helps the agent
to find better moves than what is suggested by the policy alone, it acts as an
\textit{policy improvement} operator. MCTS is integral to the process: the
strength of the trained policy is correlated with the effectiveness of MCTS as
the policy improvement operator.

Our policy is trained in generations, with the data from previous generations
used to train the next generation's policy network. The training of each
generation is synchronous, and constitutes a complete policy iteration
step. The first policy network is trained using data generated by MCTS using
the uniform random policy. Subsequent generations combine MCTS and the current
iteration of the policy to generate data. To jump-start the learning process,
the first generation is run using more simulations per step; since no policy
network is used, there is not a large run-time cost to this.

The training procedure (Algorithm~\ref{alg:training}) works as follows: We use
two queues, $B_{training}$ and $B_{validation}$, for training which store
state-action pairs $(s_t, a_t)$. As
each run finishes, the resulting pairs are stored in a temporary buffer, which
only contains data produced in the current generation. Once all runs in a
generation are finished, the pairs stored in the temporary buffer are shuffled
and split (e.g., 90-to-10), and appended to both buffers, respectively. This
ensures that both buffers will contain data from multiple generations. Note
that using a single buffer and randomly splitting it before training is not
equivalent to this procedure.

The sizes of the buffers determine the amount of data that is kept from older
generations; given more data, we can expect the policy to be better
\cite{anthony2017thinking}, but the training time will be longer. The size of
the buffers is also directly related to the amount of data generated per
run. We use the validation buffer to avoid overfitting by early stopping.

	\section{Experimental Setup}

\subsection{SameGame}

SameGame is a tile-matching game with the goal of maximizing the final score.
In each move, the player can clear horizontally or vertically connected groups of 
size two or more of equal colour. The
blocks above created holes will always fall down and then move left, if
possible. When an entire column is cleared, the columns to the right are moved
to the left. Each move scores $($\textbf{\#BlocksCleared} $- 2)^2$. The game
is over when the player cannot take an action anymore. In addition, if the
board is cleared, the player is awarded an additional $1,000$
points. Otherwise, the player receives a penalty based on the total number of colours of blocks left, which is calculated as follows: $\sum_{c}(\text{\textbf{\#BlocksLeft}}_c - 2)^2$.

Deciding whether a general SameGame instance with at least five colours and
two columns can be fully cleared has been shown to be NP-complete
\cite{schadd2008addressing}. We use boards of size $7\times 7$, $10 \times
10$, and $15 \times 15$, with five different block colours. All boards used
for training and testing are randomly generated. We slightly simplify the
action space by only allowing the agent to select the lowest left block in a
group; in the actual game, a player can click on any of the blocks in a group
to clear it.

\subsection{Policy Network Architecture}

We use the same network architecture for all of the experiments to limit its
impact on the results. The complexity of the network makes it suitable for the 15 $\times$ 15 boards,
but it is likely too deep for the simple 7 $\times$ 7 boards. 

Our architecture is similar to the one used by
\cite{anthony2017thinking}. In particular, the input to the network is an
``image'' of size $d \times d \times (c+1)$, where $d$ is the dimension of the
board and $c$ is the number of block colours, and 1 is added to encode empty
tiles. That is, the board is represented as $c+1$ binary layers to
one-hot-encode the tile state. The input is padded by 1 on all four sides,
increasing the dimension to $(d+2) \times (d+2) \times (c+1)$. This ensures
that the information at the edges of the board is not lost during
convolutions. The padded input is passed to 13 convolution layers, all of
which have 64 filters, stride of 1, and ELU activation
\cite{clevert2015fast}. Layers 11 and 13 have a kernel size of $1 \times 1$,
and all others have size $3 \times 3$.  The dimension of the input is kept for
layers 1-8 and 12, and reduced in layers 9-11 and 13. The output of the final
convolution layer is flattened and passed to a linear layer with softmax
activation, which represents the policy.

We use Adam as the optimizer, with a learning rate of
$\expnumber{5}{-4}$. Data is fed to the network in mini-batches of size
256. For regularization, early stopping on the validation loss is used.
The early stopping point is the first epoch after which the validation errors do not decrease for 3 consecutive epochs.
 We set the
size of the buffers $B_{training}$ and $B_{validation}$ to $1.5$M and $150$K,
respectively, for all experiments. Given the number of runs per generation, as provided in Table~\ref{tab:trainparams},
the buffers contain data from roughly 3 to 5 previous generations.

%
%
%
%
%

	\section{Experiments} \label{sec:exp}

In this section we will validate our contributions with several experiments in
the SameGame domain. The first compares the performance of Policy-MCTS against
plain MCTS, which is MCTS with a uniform random policy among all valid actions. 
Our experiments are run on three different board sizes using
several time budgets. To demonstrate the effectiveness of our parallelization
method, we also compare the performance of single and multi-threaded variants
of both algorithms.
To put our results into perspective, we also use Policy-MCTS to solve a
standard set of 20 SameGame test positions. These boards are commonly used to
benchmark state-of-the-art search algorithms for single-player games. We
demonstrate that a policy trained using only 25 simulations per move is
competitive with these methods.

\subsection{Comparing with Plain MCTS} \label{sec:beatMCTS}

We use our algorithm to train policy networks for board sizes $7 \times 7, 10
\times 10,$ and $15 \times 15$. The $7 \times 7$ board is small and relatively
simple. The other two sizes are increasingly more complex, with the $15 \times
15$ board often used as benchmark problem in the literature. The average
solution length of $10 \times 10$ board games is roughly twice that of $7 \times 7$,
and half of $15 \times 15$ games. To keep training time manageable, we
decrease both the number of runs and simulations as the board size increases.

\begin{table}[t]
  \centering
  \caption{Parameters used for training for each board size.}
  \label{tab:trainparams}
  \setlength{\tabcolsep}{4pt}
  \small
  \begin{tabular}{lccc}
    \toprule
    \multicolumn{1}{l}{Parameter} &
    \multicolumn{1}{c}{7 $\times$ 7}   &
    \multicolumn{1}{c}{10 $\times$ 10} &
    \multicolumn{1}{c}{15 $\times$ 15}     \\
    \midrule
    Generations		& 50		& 50		& 66 \\
    Runs/Generation	& 20,000	& 10,000	& 5,000 \\
    Simulations/Move& 100		& 50		& 25 \\
    $c_{puct}$		& 30		& 4			& 2 \\ 
    Threads/Run		& 1			& 1			& 1 \\ 
    Dirichlet Noise & 0.75		& 0.40		& 0.25 \\
    Training Time	& 3 days	& 3.5 days	& 6.5 days \\
    CPU                & Intel          & Intel    & 2 x Intel \\
    - Type           & i7-7700K & i7-8700K &  Xeon 6148 \\
    - Cores/Speed    & 4/4.2 GHz  & 6/3.7 GHz & 20/2.4 GHz \\
    GPU              & Nvidia    & Nvidia   & 4 x Nvidia  \\
    - Type/RAM       & 1070/8GB  & 1070/8GB & V100/16GB\\
  \end{tabular}
  \vspace{-0.3cm}
\end{table}

\begin{figure}[t]
  \includegraphics[width=\linewidth,height=5cm]{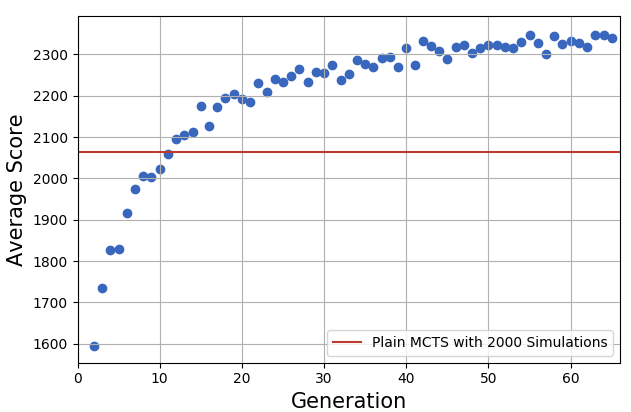}
  \caption{The average score of the 15 $\times$ 15 policy at each generation of training.}
  \label{fig:training}
  \vspace{-0.3cm}
\end{figure}

The parameters we used for training is shown in
Table~\ref{tab:trainparams}. Note that while we put effort into optimizing the
parameters, they are most likely not optimal. This is because each experiment
takes several days to run, making a grid search over these parameters not
feasible, given our computation budget. Training on 7 $\times$ 7 and 10
$\times$ 10 boards converged, but the 15 $\times$ 15 policy did not (Figure~\ref{fig:training}).
\begin{figure*}[t]
  \centering
  \includegraphics[width=0.9\linewidth,height=6cm]{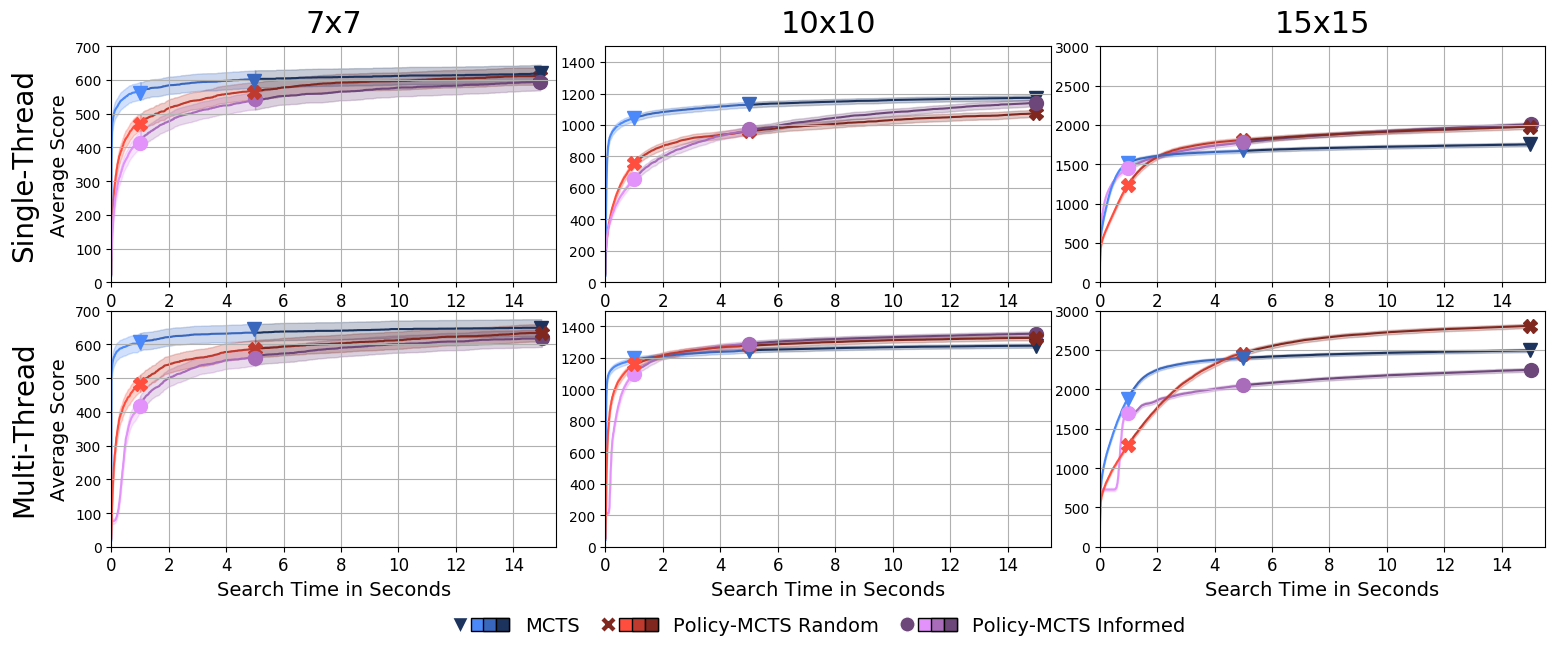}
  \caption{Average score of single and multi-threaded MCTS, Policy-MCTS with
    random rollouts, and Policy-MCTS with informed rollouts, as a function of
    search
    time in seconds. Each experiment is performed on a test set of 500 randomly
    generated boards, with 5 runs per board. Darker shades represent the same
    algorithm running for a longer period of time. The shaded area represents
    the 99\% confidence interval.}
  \label{fig:all_boards}
\end{figure*}


\begin{table*}
  \centering
  \vspace{-0.3cm}
  \setlength{\tabcolsep}{3.5pt}
  \caption{Number of simulations, node expansions, and leaf node expansions for
    single-threaded and multi-threaded experiments on the 15 $\times$ 15 boards.}
  \label{tab:alldata}
  \small
  \begin{tabular}{l|ccc|c|ccc|c|ccc}
    \multicolumn{1}{c}{}&&&\multicolumn{1}{c}{}&\multicolumn{1}{c}{}&&
    \textbf{Single-Threaded} \\
    \toprule	
    \multicolumn{1}{l|}{\textbf{Parameter}} &
    \multicolumn{3}{c|}{1 Second} & &
    \multicolumn{3}{c|}{5 Seconds} & &
    \multicolumn{3}{c}{15 Seconds}   \\ 
    \cline{2-4} \cline{6-8} \cline{10-12}
    & MCTS & P-MCTS & P-MCTS & & MCTS & P-MCTS & P-MCTS & & MCTS & P-MCTS & P-MCTS
    \\
    & & Random & Informed & & & Random & Informed & & & Random & Informed\\
    \midrule
    Simulations			& 20,106		& 1,274		& 27 	&& 127,504 & 13,748 & 131 && 403,243 & 80,580
    & 398 \\
    Expansions			& 12,074		& 1,261		& 27  	&& 27,089  & 6,128  & 131 && 46,480  & 16,426
    & 398 \\
    Leaf Expansions		& 155		& 1			& 0 	&& 809 	  & 150   & 0   && 1,731   & 628   &
    0 \\
    \toprule
    \multicolumn{1}{c}{}&&&\multicolumn{1}{c}{}&\multicolumn{1}{c}{}&&
    \textbf{Multi-Threaded} \\
    \toprule
    Simulations			& 95,335		& 9,652		& 201 	&& 697,810 & 49,538 & 864 && 2,297,200 &
    376,698 & 2,575 \\
    Expansions			& 91,140		& 9,652		& 201  	&& 310,313 & 46,295 & 864 && 572,556  &
    137,509 & 2,575 \\
    Leaf Expansions		& 148		& 0			& 0 	&& 5,522   & 110   & 0   && 15,524   & 2,691  
    & 0 \\
    \bottomrule
  \end{tabular}
  \vspace{-0.3cm}
\end{table*}

Given the same node budget, we can expect Policy-MCTS to outperform plain MCTS. 
However, using a GPU incurs a
large cost on the speed of the search. Therefore, to make a more fair
comparison, we use wall-clock time for all of our experiments. We use the 
same computer for all experiments, and optimized both MCTS and the GPU queue implementation.

Since we are using wall-clock time, we also present the averages of simulation count, terminal
node expansions, and node expansions for the experiments on the 15 $\times$ 15
boards in Table~\ref{tab:alldata}. The simulation count also contains simulations starting (and ending)
at terminal leaf nodes, i.e., terminal states that are part of the search
tree. The number of terminal node expansions is included in the node
expansions count.

We use a test set of 500 randomly generated boards for each size, to compare
Policy-MCTS with plain MCTS. We do 5 runs per board, for a total of 2,500 runs
in each experiment. Results at three different time settings of 1, 5, and 15
seconds are shown for plain MCTS, and Policy-MCTS with random and policy
guided simulations, respectively.

The same parameters are used for both the single-threaded and multi-threaded
experiments. Additionally, the $c_{puct}$ value used during training is kept
for all runs. The number of threads used by MCTS is 16, which maximizes CPU
usage, and Policy-MCTS uses 100 threads, which is sufficient to keep the GPU
fully loaded. By using such a high
number of threads, we are assuming that the extra exploration caused by
virtual loss is mitigated by the policy network.

All experiments were run using the same computer that was used to train the 10
$\times$ 10 network. The experimental results are shown in
Figure~\ref{fig:all_boards}. We observe that the multi-threaded versions of
all three algorithms outperform their single-threaded counter-parts; for the
15 $\times$ 15 board, the average score of plain MCTS at the 5 second mark
rises from 1,684 to 2,393, an increase of nearly 50\%. This provides some
evidence that our parallelization strategy is effective. Furthermore, we can
see that using multiple threads is more beneficial for both versions of
Policy-MCTS. This is expected, as using many threads allows for batch
prediction, which utilizes the GPU more effectively. We can also clearly
observe the downside of using a neural network when given a short time limit,
with plain MCTS outperforming both versions of Policy-MCTS at the 1 second
mark, in all configurations.

Another important observation is that as the decision complexity increases,
the strength difference between plain MCTS and Policy-MCTS also
increases. Given a simpler problem and a limited budget, it is probably more
beneficial to use plain MCTS. However, as the complexity of the problem
increases, so too does the benefit of using a policy. This is clearly
demonstrated in the 15 $\times$ 15 graphs, with Policy-MCTS far outperforming
plain MCTS.

A rather surprising result is the weak performance of multi-threaded
Policy-MCTS with informed rollouts, when compared to plain MCTS, on the 15
$\times$ 15 board. This leads us to speculate that given a limited time budget
on a problem with a large state and action space, creating a deeper tree by
performing many weak rollouts is better than a much smaller, more informed
number of rollouts.

Detailed performance data for these experiments is provided in
Table~\ref{tab:alldata}. For the 15 $\times$ 15 board, even with
multi-threading, only 2,575 policy guided rollouts were finished, compared to
over 2 million for plain MCTS. Note that in the single-threaded setting, 400
informed rollouts outperform over 400k random rollouts from plain MCTS. This
provides evidence that our algorithm has trained a competent policy. We can
also observe that the multi-threaded versions are far more explorative. For
example, for Policy-MCTS with random rollouts in the 5 second setting, node
expansions per simulation increases from 0.45 to 0.93.

\subsection{Comparison to State-of-the-Art Algorithms} \label{exp:compare}

In this section, we compare the performance of our parallel MCTS and
Policy-MCTS algorithms against published state-of-the-art search methods on 20
public test positions. We use the network trained on 15 $\times$ 15 boards
from the previous section, and run Policy-MCTS with random and guided rollouts, respectively; the data for the former is not presented as it performed worse than the latter. We additionally run our parallel
version of MCTS to gain a better perspective of the results obtained in the
last section. $c_{puct}$ is set to 5 and 10 and threads to 120 and 80 for Policy-MCTS and Parallel-MCTS, respectively. 
The same machinery that was used to train the 15 $\times$ 15 policy is used. All algorithms are run only once per position. 
The results of the experiment are provided in Table~\ref{tab:compare_results}.

Note that we give each algorithm only 2 hours total per position, which is
similar to SP-MCTS \cite{schadd2012single}. The results of Dist-NRPA(5)
\cite{negrevergne2017distributed} are also achieved in 2 hours, but they use
160 CPU's for each position. Algorithms NMCS(4), Sel-NRPA(4) \cite{cazenave2016nested}, 
and Div-NRPA(4)
\cite{edelkamp2016improved} take more than half a day per position; the number
in the brackets represents the nesting level used by the algorithm. Most of
these algorithms, with the exception of Div-NRPA(4) and Dist-NRPA(5), also make
use of hand-crafted heuristics to guide the rollouts.

\begin{table}[t]
\newcommand{\mcrot}[4]{\multicolumn{#1}{#2}{\rlap{\rotatebox{#3}{#4}~}}} 
\newcommand{\tilt}[1]{\mcrot{1}{l}{60}{#1}}
  \centering
  \caption{Performance of Parallel-MCTS and Policy-MCTS with guided rollouts,
  	compared to state-of-the-art search methods, on 20 public boards.}
  \label{tab:compare_results}
  \vspace{-0.3cm}
  \setlength{\tabcolsep}{2.4pt}
  \small
  \begin{tabular}{|r|r|r|r|r|r|r|r|} 
    \tilt{\small Position} & \tilt{\small SP-MCTS} & \tilt{\small NMCS(4)} &
    \tilt{\small Sel-NRPA(4)} & \tilt{\small Div-NRPA(4)} & \tilt{\small Dist-NRPA(5)} &
    \tilt{\small Parallel-MCTS} & \tilt{\small Policy-MCTS} \\ 
    \hline
    1  & 2,919 & 3,121 & 3,179 & 3,145 & \bfseries3,185 & 1,859 & 2,717 \\
    2  & 3,797 & 3,813 & \bfseries3,985 & \bfseries3,985 & \bfseries3,985 &
    3,003
    & 3,761 \\
    3  & 3,243 & 3,085 & 3,635 & \bfseries3,937 & 3,747 & 2,413 & 3,355 \\
    4  & 3,687 & 3,697 & 3,913 & 3,879 & \bfseries3,925 & 3,213 & 3,709 \\
    5  & 4,067 & 4,055 & 4,309 & 4,319 & \bfseries4,335 & 3,009 & 3,983 \\
    6  & 4,269 & 4,459 & \bfseries4,809 & 4,697 & \bfseries4,809 & 3,481 & 4,375
    \\
    7  & \bfseries2,949 & \bfseries2,949 & 2,651 & 2,795 & 2,923 & 2,473 & 2,917
    \\
    8  & 4,043 & 3,999 & 3,879 & 3,967 & 4,061 & 3,577 & \bfseries4,275 \\
    9  & 4,769 & 4,695 & 4,807 & 4,813 & 4,829 & 3,629 & \bfseries4,839 \\
    10 & \bfseries3,245 & 3,223 & 2,831 & 3,219 & 3,193 & 2,715 & 3,213 \\
    11 & 3,259 & 3,147 & 3,317 & 3,395 & \bfseries3,455 & 2,405 & 3,269 \\
    12 & 3,245 & 3,201 & 3,315 & 3,559 & \bfseries3,567 & 2,793 & 3,301 \\
    13 & 3,211 & 3,197 & 3,399 & 3,159 & \bfseries3,591 & 2,343 & 3,355 \\
    14 & 2,937 & 2,799 & 3,097 & 3,107 & \bfseries3,135 & 2,473 & 2,977 \\
    15 & 3,343 & 3,677 & 3,559 & 3,761 & \bfseries3,885 & 2,825 & 3,381 \\
    16 & 5,117 & 4,979 & 5,025 & 5,307 & \bfseries5,375 & 4,191 & 4,963 \\
    17 & 4,959 & 4,919 & 5,043 & 4,983 & \bfseries5,067 & 3,124 & 4,615 \\
    18 & 5,151 & 5,201 & 5,407 & 5,429 & \bfseries5,481 & 3,829 & 5,221 \\
    19 & 4,803 & 4,883 & 5,065 & 5,163 & \bfseries5,299 & 4,559 & 4,823 \\
    20 & 4,999 & 4,835 & 4,805 & 5,087 & \bfseries5,203 & 2,977 & 5,023 \\
    \hline
    $\Sigma$ & 78,012 & 77,934 & 80,030 & 81,706 & 83,050 & 60,891 & 78,072 \\
    \hline
  \end{tabular}
  \vspace{-0.3cm}
\end{table}
The score can potentially be increased in multiple ways. We could
evenly distribute the time budget over the average number of moves, as done in
SP-MCTS.  Additionally, an implementation capable of using multiple GPUs could
potentially achieve much higher scores with the same policy. Using a
transposition table to store predictions will also increase the speed of the
system. Smaller gains could be obtained by fine-tuning the search parameters; we
did not tune any of the parameters to these positions. Lastly, we can increase
the simulation count of MCTS during training. This greatly increases the
training time, but it also strengthens the policy evaluation and improvement
steps, which can lead to a much better policy.


\section{Conclusion and Future Work}

In this paper, we presented three main contributions. We first introduced a
novel action value normalization method, which is more general and therefore applicable 
to a wider range of domains. Then, we defined a general virtual loss function for the
single-player setting, which enabled effective tree parallelization of MCTS. Lastly, we introduced
a policy training procedure for single-agent optimization tasks. The process uses a neural network to represent the policy and MCTS
for policy improvement.

In our experiments on SameGame, we demonstrated the effectiveness of our policy training 
procedure, with the trained policy producing competitive results to state-of-the-art search algorithms on a public test set. 
Our results demonstrate a promising direction for future AI
research in single-player optimization domains. 

Another potentially fruitful research direction is to replace the policy
network with a two-headed policy and value network. The predicted value can be
combined with rollout results using a mixing parameter or replaced entirely.
Our
preliminary work on this subject has produced weaker results than plain
MCTS. Whether it is beneficial to use value prediction for single-player
optimization problems with unknown score bounds remains an open question. 
Lastly, we can look for other suitable search algorithms for policy improvement, as 
MCTS might not be the ideal choice.

	\bibliography{references}

\begin{thebibliography}{}

\bibitem[\protect\citeauthoryear{Anthony, Tian, and
  Barber}{2017}]{anthony2017thinking}
Anthony, T.; Tian, Z.; and Barber, D.
\newblock 2017.
\newblock Thinking fast and slow with deep learning and tree search.
\newblock In {\em Advances in Neural Information Processing Systems},
  5360--5370.

\bibitem[\protect\citeauthoryear{Arneson, Hayward, and
  Henderson}{2010}]{arneson2010monte}
Arneson, B.; Hayward, R.~B.; and Henderson, P.
\newblock 2010.
\newblock {M}onte {C}arlo tree search in {H}ex.
\newblock {\em IEEE Transactions on Computational Intelligence and AI in Games}
  2(4):251--258.

\bibitem[\protect\citeauthoryear{Cazenave}{2009}]{cazenave2009nested}
Cazenave, T.
\newblock 2009.
\newblock Nested {M}onte-{C}arlo search.
\newblock In {\em Twenty-First International Joint Conference on Artificial
  Intelligence}.

\bibitem[\protect\citeauthoryear{Cazenave}{2016}]{cazenave2016nested}
Cazenave, T.
\newblock 2016.
\newblock Nested rollout policy adaptation with selective policies.
\newblock In {\em Computer Games}. Springer.
\newblock  44--56.

\bibitem[\protect\citeauthoryear{Chaslot, Winands, and van
  Den~Herik}{2008}]{chaslot2008parallel}
Chaslot, G. M.-B.; Winands, M.~H.; and van Den~Herik, H.~J.
\newblock 2008.
\newblock Parallel {M}onte-{C}arlo tree search.
\newblock In {\em International Conference on Computers and Games},  60--71.
\newblock Springer.

\bibitem[\protect\citeauthoryear{Clevert, Unterthiner, and
  Hochreiter}{2015}]{clevert2015fast}
Clevert, D.-A.; Unterthiner, T.; and Hochreiter, S.
\newblock 2015.
\newblock Fast and accurate deep network learning by exponential linear units
  (elus).
\newblock {\em arXiv preprint arXiv:1511.07289}.

\bibitem[\protect\citeauthoryear{Edelkamp and
  Cazenave}{2016}]{edelkamp2016improved}
Edelkamp, S., and Cazenave, T.
\newblock 2016.
\newblock Improved diversity in nested rollout policy adaptation.
\newblock In {\em Joint German/Austrian Conference on Artificial Intelligence
  (K{\"u}nstliche Intelligenz)},  43--55.
\newblock Springer.

\bibitem[\protect\citeauthoryear{Gelly \bgroup et al\mbox.\egroup
  }{2012}]{gelly2012grand}
Gelly, S.; Kocsis, L.; Schoenauer, M.; Sebag, M.; Silver, D.; Szepesv{\'a}ri,
  C.; and Teytaud, O.
\newblock 2012.
\newblock The grand challenge of computer {G}o: {M}onte {C}arlo tree search and
  extensions.
\newblock {\em Communications of the ACM} 55(3):106--113.

\bibitem[\protect\citeauthoryear{Huang \bgroup et al\mbox.\egroup
  }{2013}]{huang2013mohex}
Huang, S.-C.; Arneson, B.; Hayward, R.~B.; M{\"u}ller, M.; and Pawlewicz, J.
\newblock 2013.
\newblock Mohex 2.0: a pattern-based mcts hex player.
\newblock In {\em International Conference on Computers and Games},  60--71.
\newblock Springer.

\bibitem[\protect\citeauthoryear{Klein}{2015}]{klein2015attacking}
Klein, S.
\newblock 2015.
\newblock Attacking samegame using {M}onte-{C}arlo tree search: using
  randomness as guidance in puzzles.

\bibitem[\protect\citeauthoryear{Kocsis and
  Szepesv{\'a}ri}{2006}]{kocsis2006bandit}
Kocsis, L., and Szepesv{\'a}ri, C.
\newblock 2006.
\newblock Bandit based {M}onte-{C}arlo planning.
\newblock In {\em European conference on machine learning},  282--293.
\newblock Springer.

\bibitem[\protect\citeauthoryear{Laterre \bgroup et al\mbox.\egroup
  }{2018}]{laterre2018ranked}
Laterre, A.; Fu, Y.; Jabri, M.~K.; Cohen, A.-S.; Kas, D.; Hajjar, K.; Dahl,
  T.~S.; Kerkeni, A.; and Beguir, K.
\newblock 2018.
\newblock Ranked reward: Enabling self-play reinforcement learning for
  combinatorial optimization.
\newblock {\em arXiv preprint arXiv:1807.01672}.

\bibitem[\protect\citeauthoryear{McAleer \bgroup et al\mbox.\egroup
  }{2018}]{mcaleer2018solving}
McAleer, S.; Agostinelli, F.; Shmakov, A.; and Baldi, P.
\newblock 2018.
\newblock Solving the {R}ubik's {C}ube without human knowledge.
\newblock {\em arXiv preprint arXiv:1805.07470}.

\bibitem[\protect\citeauthoryear{Negrevergne and
  Cazenave}{2017}]{negrevergne2017distributed}
Negrevergne, B., and Cazenave, T.
\newblock 2017.
\newblock Distributed nested rollout policy for samegame.
\newblock In {\em Workshop on Computer Games},  108--120.
\newblock Springer.

\bibitem[\protect\citeauthoryear{Pepels, Winands, and
  Lanctot}{2014}]{pepels2014real}
Pepels, T.; Winands, M.~H.; and Lanctot, M.
\newblock 2014.
\newblock Real-time {M}onte {C}arlo tree search in {M}s {P}ac-{M}an.
\newblock {\em IEEE Transactions on Computational Intelligence and AI in games}
  6(3):245--257.

\bibitem[\protect\citeauthoryear{Perez \bgroup et al\mbox.\egroup
  }{2019}]{perez2019general}
Perez, D.; Liu, J.; Abdel Samea~Khalifa, A.; Gaina, R.~D.; Togelius, J.; and
  Lucas, S.~M.
\newblock 2019.
\newblock General video game {AI}: a multi-track framework for evaluating
  agents, games and content generation algorithms.
\newblock {\em IEEE Transactions on Games}.

\bibitem[\protect\citeauthoryear{Rosin}{2011}]{rosin2011nested}
Rosin, C.~D.
\newblock 2011.
\newblock Nested rollout policy adaptation for {M}onte {C}arlo tree search.
\newblock In {\em Twenty-Second International Joint Conference on Artificial
  Intelligence}.

\bibitem[\protect\citeauthoryear{Schadd \bgroup et al\mbox.\egroup
  }{2008}]{schadd2008addressing}
Schadd, M.~P.; Winands, M.~H.; Van Den~Herik, H.~J.; and Aldewereld, H.
\newblock 2008.
\newblock Addressing {NP}-complete puzzles with {M}onte-{C}arlo methods.
\newblock In {\em Proceedings of the AISB 2008 Symposium on Logic and the
  Simulation of Interaction and Reasoning}, volume~9,  55--61.

\bibitem[\protect\citeauthoryear{Schadd \bgroup et al\mbox.\egroup
  }{2012}]{schadd2012single}
Schadd, M.~P.; Winands, M.~H.; Tak, M.~J.; and Uiterwijk, J.~W.
\newblock 2012.
\newblock Single-player {M}onte-{C}arlo tree search for samegame.
\newblock {\em Knowledge-Based Systems} 34:3--11.

\bibitem[\protect\citeauthoryear{Silver \bgroup et al\mbox.\egroup
  }{2018}]{silver2018general}
Silver, D.; Hubert, T.; Schrittwieser, J.; Antonoglou, I.; Lai, M.; Guez, A.;
  Lanctot, M.; Sifre, L.; Kumaran, D.; Graepel, T.; et~al.
\newblock 2018.
\newblock A general reinforcement learning algorithm that masters {C}hess,
  {S}hogi, and go through self-play.
\newblock {\em Science} 362(6419):1140--1144.

\bibitem[\protect\citeauthoryear{Theys \bgroup et al\mbox.\egroup
  }{2010}]{theys2010using}
Theys, C.; Br{\"a}ysy, O.; Dullaert, W.; and Raa, B.
\newblock 2010.
\newblock Using a {TSP} heuristic for routing order pickers in warehouses.
\newblock {\em European Journal of Operational Research} 200(3):755--763.

\end{thebibliography}
	\bibliographystyle{aaai}
\end{document}